\documentclass[10pt,twocolumn,letterpaper]{article}

\usepackage[pagenumbers]{cvpr} 

\newcommand{\xmark}{\ding{55}}
\newcommand{\cmarkOurs}{\textcolor{OursAccent}{\ding{51}}}

\definecolor{HeadBg}{RGB}{246,248,252}
\definecolor{Year25Row}{RGB}{255,249,230} 
\definecolor{OursRow}{RGB}{230,242,255}   
\definecolor{OursAccent}{RGB}{33,122,255} 

\newcommand{\best}[1]{\textbf{#1}}
\newcommand{\second}[1]{\underline{#1}}

\usepackage{booktabs}
\usepackage{array}
\usepackage[table]{xcolor}
\usepackage{pifont}
\usepackage{makecell}
\usepackage{fontawesome5} 

\usepackage{multirow}
\usepackage[table]{xcolor}
\usepackage{tabularx}
\usepackage{colortbl} 
\usepackage{siunitx}
\usepackage{amsmath}

\usepackage{adjustbox}
\usepackage{tcolorbox}

\usepackage{algorithm}
\usepackage{algpseudocode}
\usepackage{listings}

\definecolor{cvprblue}{rgb}{0.21,0.49,0.74}

\usepackage{glossaries}

\usepackage[pagebackref,breaklinks,colorlinks,allcolors=cvprblue]{hyperref}

\def\paperID{} 
\def\confName{CVPR}
\def\confYear{2026}

\title{Spectrally Distilled Representations Aligned with Instruction-Augmented LLMs for Satellite Imagery}

\author{Minh Kha Do$^{1}$ \quad Wei Xiang$^{1}$\footnotemark[1] \quad Kang Han$^{1}$ \quad  Di Wu$^{1}$ \quad Khoa Phan$^{1}$ \\ \quad Yi-Ping Phoebe Chen$^{1}$  \quad Gaowen Liu$^{2}$ \quad Ramana Rao Kompella$^{2}$ \\
{\normalsize $^1$La Trobe University, Melbourne, VIC 3086, Australia}\quad {\normalsize $^2$Cisco Research, San Jose, CA, USA}\\
{\tt\small \{m.do, w.xiang, k.han, d.wu, k.phan, phoebe.chen\}@latrobe.edu.au}, \\ 
{\tt\small \{gaoliu, rkompell\}@cisco.com}
}

\newglossaryentry{method}
{
    name=SATtxt,
    description={The proposed method}
}

\newglossaryentry{stage1}
{
    name=SRD,
    description={Spectral Representation Distillation}
}

\newglossaryentry{stage2}
{
    name=SGI-LLM,
    description={Spectrally-Grounded Alignment with Both Encoders Locked}
}

\newglossaryentry{dinov3txt}
{
    name=DINOv3txt,
    description={Spectrally-Grounded Alignment with Both Encoders Locked}
}

\def\llm{\text{LLM}}

\usepackage{mathtools}

\makeatletter
\usepackage{xspace}
\def\@onedot{\ifx\@let@token.\else.\null\fi\xspace}
\DeclareRobustCommand\onedot{\futurelet\@let@token\@onedot}

\def\eqref#1{equation~\ref{#1}}

\def\1{\bm{1}}

\def\eg{\emph{e.g}\onedot}

\def\ie{\emph{i.e}\onedot}

\def\etal{\emph{et al}\onedot}

\DeclareMathAlphabet{\mathsfit}{\encodingdefault}{\sfdefault}{m}{sl}
\SetMathAlphabet{\mathsfit}{bold}{\encodingdefault}{\sfdefault}{bx}{n}

\begin{document}
\maketitle
\footnotetext[1]{Corresponding author.}

\begin{abstract}
Vision-language foundation models (VLFMs) promise zero-shot and retrieval understanding for Earth observation. 
While operational satellite systems often lack full multi-spectral coverage, making RGB-only inference highly desirable for scalable deployment, the adoption of VLFMs for satellite imagery remains hindered by two factors: (1) multi-spectral inputs are informative but difficult to exploit consistently due to band redundancy and misalignment; and (2) CLIP-style text encoders limit semantic expressiveness and weaken fine-grained alignment. We present SATtxt, a spectrum-aware VLFM that operates with RGB inputs only at inference while retaining spectral cues learned during training. Our framework comprises two stages. First, Spectral Representation Distillation transfers spectral priors from a frozen multi-spectral teacher to an RGB student via a lightweight projector. Second, Spectrally Grounded Alignment with Instruction-Augmented LLMs bridges the distilled visual space and an expressive LLM embedding space. Across EuroSAT, BigEarthNet, and ForestNet, SATtxt improves zero-shot classification on average by 4.2\%, retrieval by 5.9\%, and linear probing by 2.7\% over baselines, showing an efficient path toward spectrum-aware vision-language learning for Earth observation. 
Project page: \url{https://ikhado.github.io/sattxt/}
\end{abstract}
    
\section{Introduction}
\label{sec:intro}

Vision-language foundation models (VLFMs) align visual and textual concepts through large-scale contrastive training and have reshaped representation learning~\cite{vlm_survey,FineLIP}. 
Pioneering examples include SigLIP~\cite{siglip2} and DINOtxt~\cite{dinov2txt}, which demonstrate that language supervision can drive broad generalization across modalities. 
This capability is especially valuable for satellite imagery, where exhaustive labels are rare and annotation often requires domain experts~\cite{dlsurvey, geoai_survey}. 
In this context, VLFMs make zero-shot use practical by enabling classification and retrieval, such as identifying land-cover types or rare phenomena directly from prompts~\cite{rsclip}. 
However, current VLFMs face key limitations when applied to satellite imagery, motivating the need for spectrum-aware and semantically expressive models.
\begin{figure}[t]
    \centering
    \includegraphics[width=0.8\linewidth]{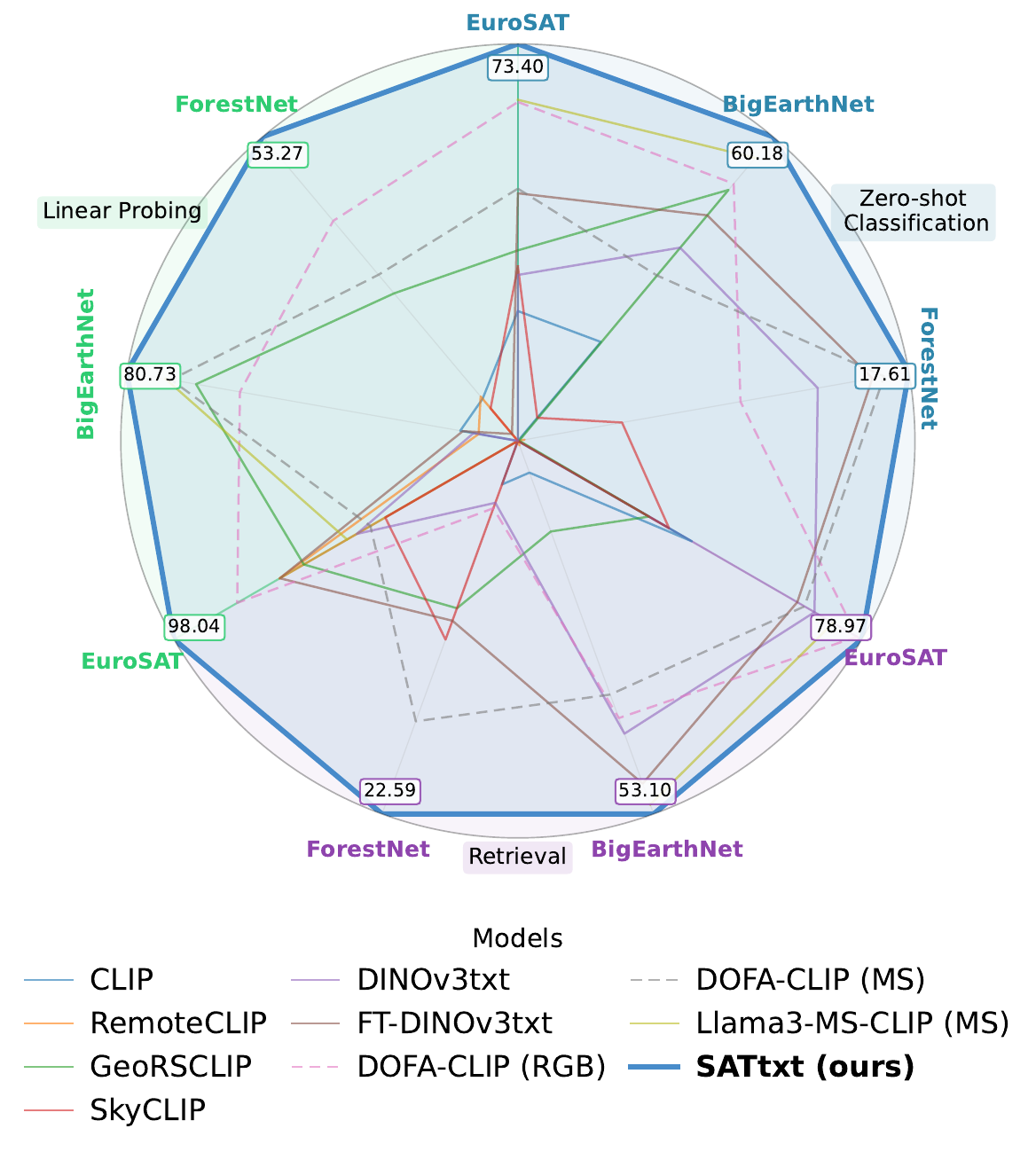}
    \caption{\gls{method} outperforms existing VLFMs across three satellite benchmarks while requiring only RGB inputs. By contrast, multi-spectral VLFMs (\eg, DOFA-CLIP~\cite{dofaclip}) exhibit inconsistent gains from multi-spectral (MS) inputs.}

    \label{fig:overall_radar_map}
\end{figure}

\begin{figure}
    \centering
    \includegraphics[width=\linewidth]{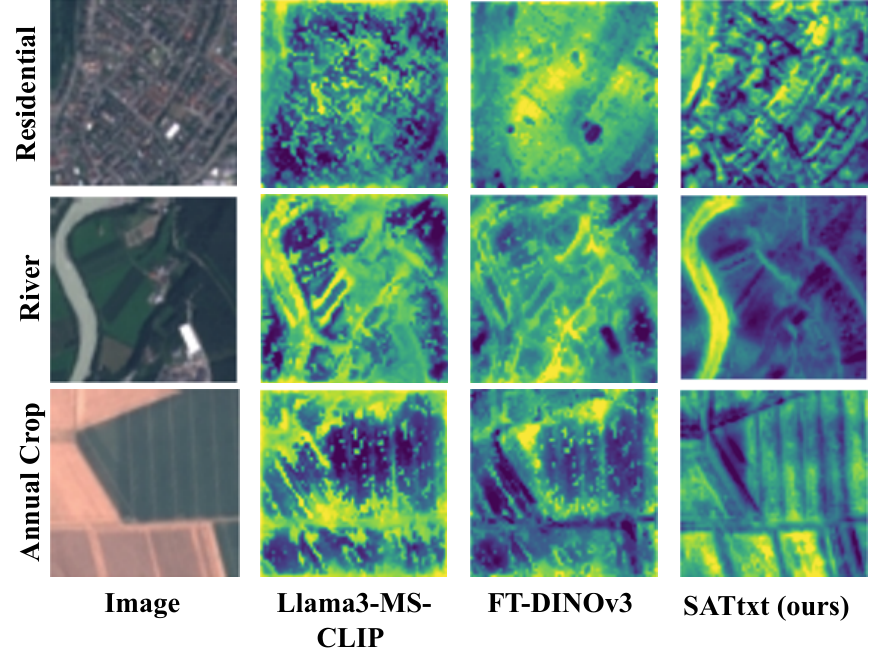}
    \caption{Patch-wise similarity between images and label prompts for models pre-trained on the same dataset. FT-\gls{dinov3txt} denotes \gls{dinov3txt} further pre-trained on this dataset, consistent with Llama3-MS-CLIP and \gls{method}. 
    Using an instruction-augmented LLM as the text encoder, our method produces sharper object-level focus (\eg, river) and clearer contextual relations (\eg, residential), supporting strong zero-shot predictions.}
    \label{fig:intro_feature_map}
\end{figure}

Meanwhile, satellites continuously collect vast amounts of unlabeled imagery from missions such as Sentinel and Landsat~\cite{EarthNets,bigearthnet,Wang2023ssl4eo}. 
However, progress in VLFMs remains constrained by the scarcity of high-quality image-text pairs~\cite{chatearthnet,earthdial}, creating a mismatch between abundant visual data and limited aligned descriptions that impedes scalable training of standard VLFMs. 
While recent efforts Llama3-MS-CLIP~\cite{ibm_ms_clip}, and DOFA-CLIP~\cite{dofaclip} report encouraging zero-shot results, two key challenges still hinder their effective deployment for satellite imagery.

First, satellite imagery commonly comprises multi-spectral inputs, yet existing models often fail to exploit them effectively. 
While additional bands provide complementary information, they also introduce redundancy and inter-band misalignment
~\cite{band_missalignment,ibm_ms_clip}.
Empirical studies~\cite{ibm_ms_clip} show that adding spectral bands yields diminishing or unstable gains.
For instance,~\cref{fig:overall_radar_map} illustrates DOFA-CLIP~\cite{dofaclip} achieving fluctuating gains.
In addition, Llama3-MS-CLIP~\cite{ibm_ms_clip} observes degradation beyond ten bands. 
Moreover, complete spectral stacks are not always available due to atmospheric conditions or sensor degradation~\cite{bandreconstruction}. 
These factors motivate the development of an RGB-only VLFM that exploits prior spectral knowledge while avoiding the redundancy and misalignment of explicit multi-spectral stacks. We show that such a model can often match or exceed the performance of MS models on standard benchmarks.

Second, recent VLFMs for satellite imagery remain constrained by the limited representational capacity of the CLIP-style text encoder. 
Prior approaches such as RemoteCLIP~\cite{remoteclip}, GeoRSCLIP~\cite{GeoRSCLIP}, and Llama3-MS-CLIP~\cite{ibm_ms_clip} continue to pre-train or extend both CLIP encoders.
Other approaches, such as DOFA-CLIP~\cite{dofaclip}, which freezes the text encoder, and DINOv3txt, which adopts a LiT-like design~\cite{lit} that freezes the vision backbone, still depend on a comparatively lightweight text encoder.
Across these variants, the text encoder remains the persistent bottleneck, restricting textual expressivity and weakening fine-grained cross-modal alignment. 
In contrast, advances in large language models (LLMs) demonstrate that instruction-augmented language representations can produce more expressive embeddings~\cite{llm2vec,skean2025layer,William2026}. 
Building on this insight, we employ an instruction-augmented LLM to generate dense text embeddings that replace standard CLIP text features. 
As illustrated in~\cref{fig:intro_feature_map}, these LLM-derived embeddings enhance object-level focus and contextual reasoning, leading to consistent gains in downstream tasks.\\
\indent
To address the aforementioned challenges, we propose \gls{method}, an RGB-only VLFM for satellite imagery that retains spectral cues and rich language grounding. 
Our model is pre-trained in two stages. 
First, \textbf{Spectral Representation Distillation} (\gls{stage1}) transfers knowledge from a multi-spectral encoder to an RGB encoder via a lightweight projector.
The projector reconstructs multi-spectral representations from RGB representations, enabling the model to effectively exploit spectral knowledge.
Second, \textbf{Spectrally Grounded Alignment with Instruction-Augmented LLMs} (\gls{stage2}) aligns the spectrally distilled representations with instruction-augmented LLMs via lightweight projectors trained under a contrastive learning objective. 
By exploiting the richer semantics of instruction-augmented LLMs, this stage overcomes the limitations of CLIP-style text encoders and yields stronger cross-modal representations.
This frozen-backbone design substantially reduces pre-training cost.\\
\indent
In summary, our contributions are threefold. 
(1) We propose SATtxt, a spectrum-aware VLFM for satellite imagery that leverages spectral priors while operating exclusively on RGB inputs at inference. 
(2) We propose Spectral Representation Distillation (SRD), which transfers multi-spectral priors into an RGB-based representation space, enabling spectrum-aware reasoning without multi-spectral inputs during inference. 
(3) We design Spectrally Grounded Alignment with Instruction-Augmented LLMs (SGI-LLM), an alignment stage that bridges spectrally distilled visual representations into the space of instruction-augmented LLM embeddings via lightweight projectors, thereby producing spectrally grounded and semantically expressive cross-modal representations.
Extensive experiments across four benchmarks,~\ie, EuroSAT,  BigEarthNet, ForestNet, and DFC2020
demonstrate that SATtxt outperforms multi-spectral baselines in zero-shot classification, text-image retrieval, open vocabulary segmentation and linear probing.
\section{Related Work}
\label{sec:related_work}
\begin{figure}
    \centering
    \includegraphics[width=\linewidth]{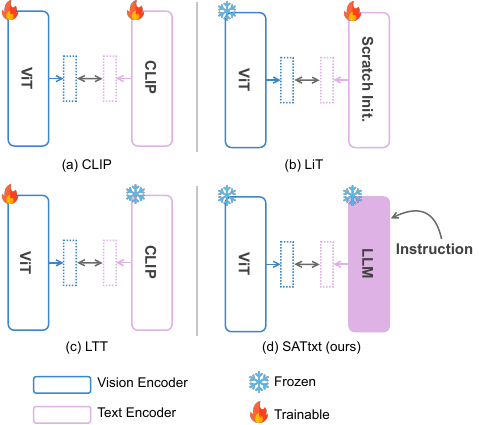}
\caption{Pre-training strategies for VLFMs on satellite imagery. (a) CLIP-style continued pre-training (\eg, RemoteCLIP), (b) LiT: Locked-image Tuning (\eg, \gls{dinov3txt}), (c) LTT: Locked-text Tuning (\eg, DOFA-CLIP), and (d) \gls{method}: bridges two strong, frozen encoders via instruction-augmented text from an LLM, improving training efficiency as well as zero-shot and linear-probe performance.}
    \label{fig:freeze_both}
\end{figure}

\subsection{Representation Learning in Remote Sensing}

Remote sensing (RS) foundation models aim to support diverse downstream tasks with limited annotation and minimal adaptation~\cite{geoai_survey,li2025masked}. Most existing RS pre-training pipelines follow self-supervised learning (SSL), particularly masked image modeling (MIM)~\cite{MIM,RingMo,robsense} variants such as SatMAE~\cite{satmae2022,satmaepp}, Scale-MAE~\cite{scaleMAE}, and SpectralGPT~\cite{spectralGPT}, or multi-modal extensions such as CROMA~\cite{croma}, Skysense~\cite{skysense}, and Terramind~\cite{terramind_iccv_2025}. While reconstruction  objectives effectively learn spatial and spectral priors, they tend to emphasize low-level statistics, often requiring heavy task-specific tuning to obtain high-level semantic capability~\cite{JEPA}.
Beyond MIM, self-distillation approaches~\cite{anysat, dinov1} match representations across augmented views, obviating the need for explicit part reconstruction within the JEPA framework~\cite{JEPA} and obtaining strong performance.
However, they typically require teacher and student to share architecture and modality, provide no mechanism to inject language semantics or spectral cues.
This limits their ability to generalize to heterogeneous inputs and open-vocabulary settings.

We introduce~\gls{stage1}, a cross-modal knowledge distillation framework that transfers spectral priors from MS imagery to RGB models.
While various forms of knowledge distillation exist for remote sensing~\cite{cls_distill, distill_survey, skysense},
the key idea in SRD is a cross-modal, unidirectional distillation from an MS teacher to an RGB student through a lightweight projector. 
This mechanism is specifically designed to imbue an RGB encoder with spectral knowledge without requiring MS inputs at inference.
Unlike existing methods~\cite{dinov1, oquab2023dinov2}, the two operate on different modalities and may use distinct, modality-specific backbones.
Moreover, both encoders stay frozen. 
Consequently, a lightweight projector is the only trainable module, aligning RGB features to the teacher’s MS space and thus transferring spectral knowledge for efficient adaptation without full fine-tuning.

\subsection{Vision-Language Foundation Models for Remote Sensing}

Contrastive learning VLFMs demonstrate strong zero-shot transfer by aligning images and text in a shared embedding space~\cite{ClearCLIP, siglip2}. 
In remote sensing, representative adaptations include RemoteCLIP~\cite{remoteclip}, SkyCLIP~\cite{skyclip}, and GeoRSCLIP~\cite{GeoRSCLIP}, which are trained on large-scale RS RGB datasets whose textual descriptions are primarily derived from labeled categories or short annotations. 
While effective, such captions are brief and semantically shallow, which can limit generalization.
More recently, Llama3-MS-CLIP~\cite{ibm_ms_clip} extends alignment to multi-spectral (MS) imagery by pairing MS images with rich and detailed descriptions, and DOFA-CLIP~\cite{dofaclip} introduces modality-aware knowledge aggregation together with wavelength-based patch embeddings~\cite{dofa} to accommodate varying spectral bands. 
Despite these advances, current RS VLFMs still face challenges in fully exploiting multi-spectral inputs, and their representation richness can be constrained by conventional text encoders.

\cref{fig:freeze_both} illustrates that prior work typically employs either a CLIP text encoder or a text encoder trained from scratch (scratch init.) and fine-tunes at least one encoder.
By contrast, SATtxt follows a different learning pattern.
Specifically, we freeze both the RGB vision encoder and the LLM-based text encoder, and train only lightweight projectors. 
Spectral knowledge is distilled via cross-modal distillation from the MS teacher into the frozen RGB encoder, and cross-modal alignment is achieved by connecting this spectrally informed RGB space to a frozen instruction-augmented LLM. 
This design yields RGB-only inference and stronger zero-shot transfer with a low pre-training cost.

\section{Method}
\label{sec:method}

\begin{figure*}[t]
    \centering
    \includegraphics[width=\linewidth]{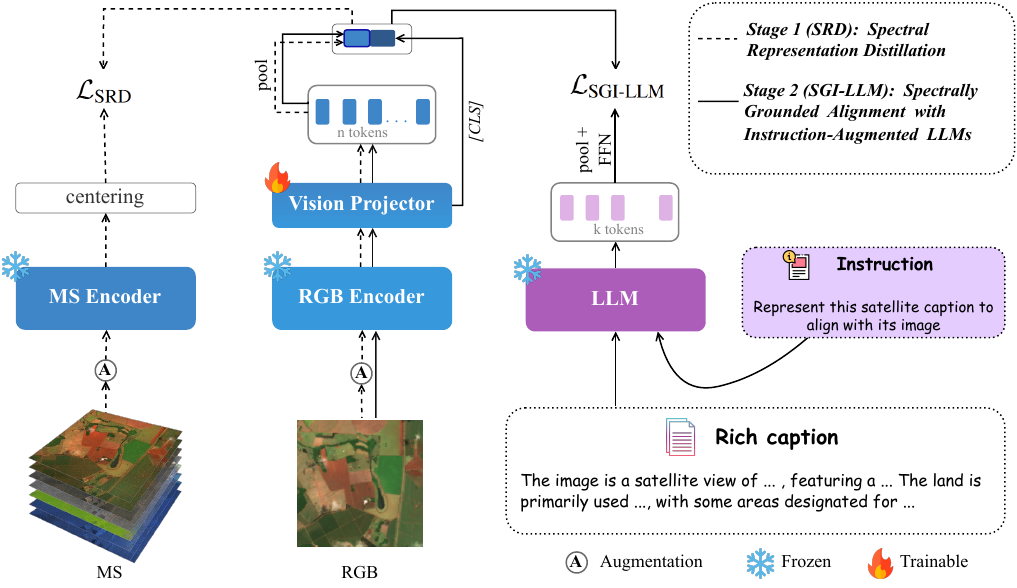}
    \caption{Two-stage pre-training pipeline for \gls{method}. 
    \textbf{Stage~1 (\gls{stage1})} - dashed lines: a vision projector is trained to reconstruct multi-spectral representations from an RGB encoder by distilling a frozen MS teacher, transferring spectral knowledge so MS inputs are unnecessary in Stage~2 and at inference. 
    \textbf{Stage~2 (\gls{stage2})} - solid lines: with vision and text encoders frozen, distilled vision features are aligned with LLM-based text embeddings using instruction-augmented prompts, enhancing cross-modal representations while preserving pretrained capabilities.}
    \label{fig:overall_pipeline}
\end{figure*}

In this section, we detail \gls{method}. 
\cref{fig:overall_pipeline} illustrates the two pre-training stages. 

\subsection{Spectral Representation Distillation}
Multi-spectral imagery includes both visible (RGB) and non-visible bands, providing richer spectral cues than RGB alone. 
However, directly relying on MS inputs yields inconsistent performance gains~\cite{ibm_ms_clip} and reduces deployability at test time.
To bridge this gap, we train a lightweight projector to reconstruct MS representations from RGB features, transferring spectral priors to the RGB encoder. 
This enables spectrum-aware representations while requiring only RGB inputs during inference.
The overall training procedure is summarized in~\cref{algo:SRD}.

Given an MS image $\mathbf{x}_{\mathrm{ms}}$, we construct two augmented view sets: an MS set $\mathcal{V}_{\mathrm{ms}}$ and an RGB set $\mathcal{V}_{\mathrm{rgb}}$. 
The MS set contains full-resolution, full-band \emph{global} views and is encoded by a frozen, pretrained MS encoder $\mathcal{E}_{\mathrm{ms}}$ to produce teacher pre-softmax outputs. 
The RGB set follows a multi-crop strategy with local crops $\mathcal{V}^{\mathrm{local}}_{\mathrm{rgb}}$ and full-resolution global views $\mathcal{V}^{\mathrm{global}}_{\mathrm{rgb}}$, such that $\mathcal{V}_{\mathrm{rgb}}=\mathcal{V}^{\mathrm{local}}_{\mathrm{rgb}}\cup\mathcal{V}^{\mathrm{global}}_{\mathrm{rgb}}$. 
These RGB views are encoded by a frozen, pretrained RGB encoder $\mathcal{E}_{\mathrm{rgb}}$. 
A lightweight projector $\mathcal{G}_v: \mathbb{R}^{d_{\mathrm{rgb}}}\!\to\!\mathbb{R}^{K}$ is the only trainable module, mapping RGB representations into the MS representation space of dimension $K$, where $K$ denotes the patch embedding dimensionality of the MS encoder. 
For each RGB view $v \in \mathcal{V}_{\mathrm{rgb}}$ and MS view $u \in \mathcal{V}_{\mathrm{ms}}$,
\begin{align}
\mathbf{z}^{\mathrm{rgb}}_{v} &= \mathcal{E}_{\mathrm{rgb}}\!\big(\tilde{\mathbf{x}}^{(v)}_{\mathrm{rgb}}\big) \in \mathbb{R}^{d_{\mathrm{rgb}}},\\
\mathbf{z}^{\mathrm{ms}}_{u}  &= \operatorname{Pool}\!\left(\mathcal{E}_{\mathrm{ms}}\!\big(\tilde{\mathbf{x}}_{\mathrm{ms}}^{(u)}\big)\right) \in \mathbb{R}^{K}, \\
\hat{\mathbf{z}}^{\mathrm{ms}}_{v} &= \mathcal{G}_v\!\big(\mathbf{z}^{\mathrm{rgb}}_{v}\big) \in \mathbb{R}^{K},
\end{align}
where tildes denote augmented views and $\operatorname{Pool}(\cdot)$ is patch-wise average pooling over MS tokens. 
Intuitively, the projector $\mathcal{G}_v$ learns to map RGB features into the MS teacher’s embedding space, allowing each RGB view to be represented as if it had been observed in multi-spectral bands. 
The resulting RGB encoder, equipped with distilled spectral priors, serves as initialization for the subsequent LLM-alignment stage.

Training adopts the centering and temperature sharpening strategy of DINO~\cite{dinov1}.
However, in DINO, the teacher network is updated via exponential moving average of the student, whereas here the teacher is a frozen, pretrained MS encoder. 
Let $\boldsymbol{\mu}\!\in\!\mathbb{R}^{K}$ denote an exponential moving average of the teacher's pre-softmax outputs with momentum $m_c$, 
and let $\tau_t < \tau_s$ be the temperature parameters for the teacher and student, respectively. 
We define
\begin{align}
\mathbf{q}_{u} &= \mathrm{softmax}\!\Big(\tfrac{\mathbf{z}^{\mathrm{ms}}_{u} - \boldsymbol{\mu}}{\tau_t}\Big), \qquad
\mathbf{p}_{v} = \mathrm{softmax}\!\Big(\tfrac{\hat{\mathbf{z}}^{\mathrm{ms}}_{v}}{\tau_s}\Big),
\end{align}
and minimize the average cross-entropy over all RGB-MS view pairs
\begin{equation}
\label{eq:srd-loss}
\mathcal{L}_{\text{SRD}}
=
\frac{1}{|\mathcal{V}_{\mathrm{rgb}}|\,|\mathcal{V}_{\mathrm{ms}}|}
\sum_{v\in\mathcal{V}_{\mathrm{rgb}}}\sum_{u\in\mathcal{V}_{\mathrm{ms}}}
\big[-\, \mathbf{q}_{u}^\top \log \mathbf{p}_{v} \big].
\end{equation}
This objective encourages each RGB view to match the distribution of its MS teacher, 
thereby transferring spectral priors into the RGB encoder.

Only the projector $\mathcal{G}_v$ is updated during training, while both $\mathcal{E}_{\mathrm{rgb}}$ and $\mathcal{E}_{\mathrm{ms}}$ remain frozen. 
Let $\operatorname{sg}(\cdot)$ denote the stop-gradient operator, defined as $\operatorname{sg}(\mathbf{z}) = \mathbf{z}$ in the forward pass and $\partial\,\operatorname{sg}(\mathbf{z})/\partial \mathbf{z} = 0$ during backpropagation. 
The center update is given by
\begin{equation}
\boldsymbol{\mu} \leftarrow m_c\,\boldsymbol{\mu} + (1-m_c)\,\frac{1}{|\mathcal{V}_{\mathrm{ms}}|}\sum_{u\in\mathcal{V}_{\mathrm{ms}}}\operatorname{sg}\!\big(\mathbf{z}^{\mathrm{ms}}_{u}\big).
\end{equation}
After distillation, the projector $\mathcal{G}_v$ learns to reconstruct MS representations from RGB input alone, 
allowing all subsequent stages and inference to operate solely on RGB inputs.

\begin{algorithm}[tb]
\caption{Spectral Representation Distillation (PyTorch-like pseudocode; multi-crop omitted for clarity)}
\label{algo:SRD}
\definecolor{codeblue}{rgb}{0.25,0.5,0.5}
\lstset{
  basicstyle=\fontsize{7.2pt}{7.2pt}\ttfamily\bfseries,
  commentstyle=\fontsize{7.2pt}{7.2pt}\color{codeblue},
  keywordstyle=\fontsize{7.2pt}{7.2pt},
}
\begin{lstlisting}[language=python]
# e_rgb, e_ms: frozen RGB and MS encoders
# g: trainable projector mapping RGB to MS space
# C: EMA center in R^K
# T_t < T_s: temperatures; m: center momentum

for x_ms, x_rgb in loader:  # paired MS/RGB minibatch
    u, v = aug_ms(x_ms), aug_rgb(x_rgb)
    
    t = e_ms(u)             # teacher logits in R^K
    s = g(e_rgb(v))         # student logits in R^K
    
    q = softmax((t - C) / T_t, dim=1)
    p = softmax(s / T_s, dim=1)
    
    loss = -(q * log(p)).sum(dim=1).mean()
    loss.backward()
    update(g)               # update projector only

    with torch.no_grad():   # EMA center update
        C = m*C + (1 - m)*t.mean(dim=0)
\end{lstlisting}
\end{algorithm}

Our method extends DINO’s centering and temperature sharpening with three adaptations for spectral-to-RGB transfer. 
First, the teacher is a fixed MS encoder, providing a stable spectral reference and preventing drift when RGB features must capture information beyond their spectral range. 
Second, both encoders are frozen and only the lightweight projector is trained, concentrating capacity on cross-modal mapping while reducing overfitting and computational cost. 
Third, we perform cross-modal alignment by pairing RGB views with MS teacher outputs, using multi-crop augmentations for RGB and global views for MS.
This asymmetric design enhances RGB diversity while maintaining a coherent spectral target.
MS inputs are used only once during SRD to distill spectral knowledge.
Stage~2 and downstream tasks operate only on RGB imagery.
\subsection{Spectrally Grounded Alignment with Instruction-Augmented LLMs}
In contrastive learning, the expressive capacity of modality encoders is crucial. 
Many remote-sensing VLFMs still rely on conventional text encoders, which limits semantic richness. 
Recent studies demonstrate that LLMs serve not only as strong generators but also as competitive text encoders~\cite{llm2vec,skean2025layer}. 
Unlike LiT~\cite{lit}, which freezes a powerful image encoder while updating the text encoder, 
we employ an LLM as the text encoder, freeze both encoders, and train only lightweight projectors for alignment. 
This design preserves pretrained capabilities while yielding richer and more expressive cross-modal representations.

Given an RGB image $\mathbf{x}_{\mathrm{rgb}}$ and a composite textual prompt consisting of a caption $\mathcal{C}$ and an instruction $\mathcal{I}$, 
the frozen image encoder and learned projector produce 
$\mathbf{H}_v = \mathcal{G}_v\!\big(\mathcal{E}_{\mathrm{rgb}}(\mathbf{x}_{\mathrm{rgb}})\big) \in \mathbb{R}^{(1+n)\times d_v}$, 
while the LLM-based text encoder generates 
$\mathbf{H}_t = \llm(\mathcal{C},\mathcal{I}) \in \mathbb{R}^{k\times d_t}$. 
Here, $\mathbf{H}_v$ contains one class token and $n$ patch tokens of width $d_v$, 
and $\mathbf{H}_t$ consists of $k$ token-level embeddings of width $d_t$. 
Following DINOtxt~\cite{dinov2txt,dinov3}, we concatenate the class token with the mean of patch tokens to form a compact and robust visual descriptor $\mathbf{z}_v$ as
\begin{equation}
\label{eq:visual_feature}
\mathbf{z}_v
=
\big[\, \mathbf{H}_v^{\langle \mathrm{cls}\rangle};
\operatorname{mean}\!\big(\mathbf{H}_v^{\text{patch}}\big)
\,\big]
\in
\mathbb{R}^{2d_v}.
\end{equation}

The frozen LLM encodes the instruction-augmented prompt $(\mathcal{C}, \mathcal{I})$. 
Mean pooling over all token embeddings produces a sentence-level representation $\tilde{\mathbf{z}}_t$, 
which is subsequently projected into the shared vision-language space 
\begin{equation}
\label{eq:text_feature}
\tilde{\mathbf{z}}_t = \operatorname{mean}(\mathbf{H}_t) \in \mathbb{R}^{d_t},
\qquad
\mathbf{z}_t = \mathcal{G}_t\!\big(\tilde{\mathbf{z}}_t\big) \in \mathbb{R}^{2d_v},
\end{equation}
where $\mathcal{G}_t$ is a lightweight linear projector. 
This projection aligns the LLM-derived textual embedding with the visual descriptor $\mathbf{z}_v$ in a unified feature space.

Using $\llm$ as the text encoder offers two key advantages. 
First, since $\llm$ is frozen, $\mathbf{H}_t$ can be precomputed and cached once per prompt, significantly reducing training time. 
Second, unlike traditional text encoders with limited token budgets (\eg, 77 tokens in CLIP~\cite{openaiclip}), 
the LLM encoder supports longer instruction-augmented inputs $(\mathcal{C}, \mathcal{I})$, 
providing richer semantics and task-aware signals that strengthen cross-modal alignment.

\noindent
\textbf{Contrastive objective.} We align the visual and textual embeddings $\mathbf{z}_v$ and $\mathbf{z}_t$ using a symmetric InfoNCE objective over a mini-batch $\mathcal{B}$. 
Let $s(\cdot,\cdot)$ denote cosine similarity and $\tau$ the temperature parameter. 
The image-to-text and text-to-image contrastive losses are defined as
\begin{equation}
\label{eq:contrastive_loss}
\begin{split}
\mathcal{L}_{v\rightarrow t}
&= -\frac{1}{|\mathcal{B}|}\sum_{i\in\mathcal{B}}
\log
\frac{\exp\big(s(\mathbf{z}_{v,i},\mathbf{z}_{t,i})/\tau\big)}
{\sum_{j\in\mathcal{B}}\exp\big(s(\mathbf{z}_{v,i},\mathbf{z}_{t,j})/\tau\big)},\\
\mathcal{L}_{t\rightarrow v}
&= -\frac{1}{|\mathcal{B}|}\sum_{i\in\mathcal{B}}
\log
\frac{\exp\big(s(\mathbf{z}_{t,i},\mathbf{z}_{v,i})/\tau\big)}
{\sum_{j\in\mathcal{B}}\exp\big(s(\mathbf{z}_{t,i},\mathbf{z}_{v,j})/\tau\big)}.
\end{split}
\end{equation}
The final training objective averages both directions
\begin{equation}
\mathcal{L}_{\text{\gls{stage2}}}
=
\frac{1}{2}\Big(\mathcal{L}_{v\rightarrow t} + \mathcal{L}_{t\rightarrow v}\Big).
\end{equation}
This symmetric formulation ensures bidirectional consistency between the visual and textual modalities.

At inference time, textual embeddings for the label set can be precomputed and cached once, 
so that the per-query cost is dominated by a single forward pass through the RGB vision encoder and the projectors. 
Consequently, employing a large LLM-based text encoder does not significantly impact online latency.

\section{Experiments}
\label{sec:experiments}


\textbf{Pre-training.}
We use SSL4EOS12~\cite{ssl4eos12_2025}, a widely used dataset for geospatial foundation models~\cite{ibm_ms_clip, terramind_iccv_2025}, as the training set for both stages.
The dataset comprises approximately $1$\,M images with 12 spectral bands and global geographic coverage. 
The image-caption pairs are from the public Llama3-SSL4EO-S12 v1.1 caption set~\cite{ibm_ms_clip}, the same dataset used by Llama3-MS-CLIP.
The RGB encoder is the distilled ViT-L variant of DINOv3~\cite{dinov3} pretrained on satellite imagery.
In the first stage, the MS teacher is SpectralGPT~\cite{spectralGPT}, a pre-trained MIM model for MS data. 
The vision projector remains lightweight, mirroring the DINOtxt configuration with two transformer blocks. 
In the second stage, the text encoder is Llama-3.1-8B initialized from LLM2Vec~\cite{llm2vec}. 
We hold both encoders fixed, pre-compute and cache the text embeddings.
All stages are conducted on $8\times$NVIDIA H200 GPUs (141 GB each). Stage 1 runs for $\sim$4  hours and Stage 2 for $\sim$3 hours, for a total of  $\sim$7 hours of pre-training.

\noindent
\textbf{Downstream datasets.}
We evaluate on three unseen satellite benchmarks: EuroSAT~\cite{eurosat}, BigEarthNet~\cite{bigearthnet}, and ForestNet~\cite{forestnet}. 
EuroSAT and BigEarthNet are derived from Sentinel-2 imagery. The dataset details can be found in the Supplementary Material.

\noindent
\textbf{Baselines.}
We evaluate the proposed method against VLFMs in both general-purpose and remote sensing domains.
The models and their vision-input types are summarized in~\cref{tab:zscls_retrieval}. 
While most baselines accept only RGB inputs, Llama3-MS-CLIP~\cite{ibm_ms_clip} and DOFA-CLIP~\cite{dofaclip} are early models that support MS inputs. 
The released Llama3-MS-CLIP weights handle only 10 spectral bands, whereas DOFA-CLIP~\cite{dofaclip} accommodates variable band configurations.
Consequently, it is not applicable to the ForestNet dataset (5 bands), and we mark the corresponding table entries with ``-''.
For fairness, we also fine-tune \gls{dinov3txt} on the same training data as our method, denoted FT-\gls{dinov3txt}.
Consequently, the \gls{method}, Llama3-MS-CLIP, and FT-\gls{dinov3txt} models are all trained on the same SSL4EO-S12 dataset.

\begin{table*}[t]
\centering
\caption{Zero-shot classification and text-to-image retrieval on three satellite benchmarks. The “Locked?~\faLock” column indicates whether the text/vision encoders are frozen or adapter-tuned (\cmarkOurs/LoRA) vs.\ fully unfrozen (\xmark). “-” denotes an unsupported task or an inapplicable dataset.}

\label{tab:zscls_retrieval}
\footnotesize 
\sisetup{detect-weight, mode=text} 
\setlength{\tabcolsep}{3pt} 

\begin{tabular}{
    >{\raggedright\arraybackslash}p{1.6cm} 
    l 
    c 
    c 
    c 
    S[table-format=2.2] 
    S[table-format=2.2] 
    S[table-format=2.2] 
    S[table-format=2.2] 
    S[table-format=2.2] 
    S[table-format=2.2] 
}
\toprule
\multirow{2}{*}{\makecell[l]{\textbf{Training}\\\textbf{Approach}}} & \multirow{2}{*}{\textbf{Model}} & \multirow{2}{*}{\textbf{Input}} & \multicolumn{2}{c}{\textbf{Locked?} \, \faLock} & \multicolumn{3}{c}{\textbf{Zero-shot Classification}} & \multicolumn{3}{c}{\textbf{Retrieval}} \\
\cmidrule(lr){4-5} \cmidrule(lr){6-8} \cmidrule(lr){9-11}
& & & \textbf{Text} & \textbf{Vision} & {\textbf{EuroSAT}} & {\textbf{BigEarthNet}} & {\textbf{ForestNet}} & {\textbf{EuroSAT}} & {\textbf{BigEarthNet}} & {\textbf{ForestNet}} \\
\midrule
\multirow{2}{*}{Generative}
& GeoChat~\cite{geochat} & RGB & LoRA & \xmark & 35.00 & 24.16 & 6.31 & {-} & {-} & {-} \\
& EarthDial~\cite{earthdial} & MS & LoRA & \xmark  & 63.28 &52.60 & 8.22 & {-} & {-} & {-} \\
\midrule 
\multirow{10}{*}{\makecell[l]{Contrastive\\}}
& CLIP~\cite{openaiclip} & RGB & \xmark  & \xmark & 46.90 & 54.85 & 8.30 & 56.92 & 28.57 & 11.78 \\
& RemoteCLIP~\cite{remoteclip} & RGB & \xmark  & \xmark & 34.02 & 52.28 & 8.46 & 34.34 & 26.62 & 10.35 \\
& GeoRSCLIP~\cite{GeoRSCLIP} & RGB & \xmark  & \xmark & 52.92 & 58.80 & 8.33 & 51.36 & 32.80 & 15.84 \\
& SkyCLIP~\cite{skyclip} & RGB & \cmarkOurs & \xmark  & 51.33 & 52.88 & 10.78 & 53.96 & 26.29 & 16.87 \\
& \gls{dinov3txt}~\cite{dinov3} & RGB & \xmark  & \cmarkOurs & 50.48 & 57.30 & 15.44 & 72.84 & 47.33 & 12.38 \\
& FT-\gls{dinov3txt} & RGB & \xmark  & \cmarkOurs & 58.58 & 58.14 & 16.74 & 70.60 & 50.92 & 16.25 \\
& DOFA-CLIP~\cite{dofaclip} & RGB & \cmarkOurs & \xmark & 67.64 & 58.96 & 13.60 & \second{78.36} & 46.21 & 12.56 \\
& DOFA-CLIP~\cite{dofaclip} & MS & \cmarkOurs  & \xmark & 59.04 & 56.58 & \second{17.02} & 71.54 & 44.52 & \second{19.55} \\
& Llama3-MS-CLIP~\cite{ibm_ms_clip} & MS & \xmark & \xmark  & \second{67.86} & \second{59.63} & {-} & 75.26 & \second{52.42} & {-} \\
\cmidrule(lr){2-11}
& \textbf{\gls{method} (ours)} & RGB & \cmarkOurs & \cmarkOurs &  \best{73.40} & \best{60.18} & \best{17.61} & \best{78.97} & \best{53.10} & \best{22.59} \\
\bottomrule
\end{tabular}
\end{table*}

\noindent
\textbf{Downstream tasks.}
We evaluate models on four fundamental downstream tasks,~\ie, zero-shot classification, text-to-image retrieval, open-vocabulary segmentation, and linear probing.
For zero-shot classification and text-to-image retrieval, we follow the evaluation protocol of Clive~\etal~\cite{ibm_ms_clip}.
For zero-shot classification, we adopt the prompt template from GeoRSCLIP~\cite{GeoRSCLIP} for each class.
For BigEarthNet, we adopt the setup of Clive~\etal~\cite{ibm_ms_clip}, treating the task as multi-label and decomposing it into $K$ one-vs-rest binary classifiers. 
We therefore report accuracy following the Llama3-MS-CLIP setup for fair comparison, while F1 scores are provided in the Supplementary Material.
For text-to-image retrieval, we compute the similarity between a given text label and all test images, rank the scores in descending order, and compute the mean average precision over the top 100 results. 
We report the class-averaged mAP@100.
In addition, for the open vocabulary, we follow the decoder-free approach from DINOtxt~\cite{dinov2txt}. For this task, we use DFC2020 dataset with Sentinel-2 images.

For linear probing, we follow the PANGAEA setup~\cite{pangaea}.
We extract final-layer representations from the models and train a single linear classifier on top.
On BigEarthNet, due to the scale of the dataset, we additionally evaluate using only 10\% of the training set following~\cite{satmae2022, SeasonalContrast}.
We report mAP~$\uparrow$ on BigEarthNet and accuracy on the other datasets. 
Additionally, F1 scores and detailed implementations are provided in the Supplementary Material.

\subsection{Results}
\noindent
\textbf{Zero-shot classification \& retrieval.}
We evaluate against both contrastive and generative multi-modal baselines, including GeoChat~\cite{geochat} and EarthDial~\cite{earthdial} for classification.

\cref{tab:zscls_retrieval} reports results on EuroSAT, BigEarthNet, and ForestNet. 
Overall, generative baselines show limited transfer to remote-sensing categories, whereas contrastive approaches yield stronger cross-modal alignment. 
Domain-adapted variants of CLIP improve over general-purpose CLIP, underscoring the benefit of remote sensing aware pre-training.  
Our \gls{method} attains the strongest zero-shot performance across datasets. 
It consistently outperforms the best prior contrastive baselines.
Notably, \gls{method} retains these gains even when tested with RGB inputs, suggesting that its spectral-aware training yields a representation that transfers beyond the training sensor configuration.

\begin{figure}
    \centering
    \includegraphics[width=\linewidth]{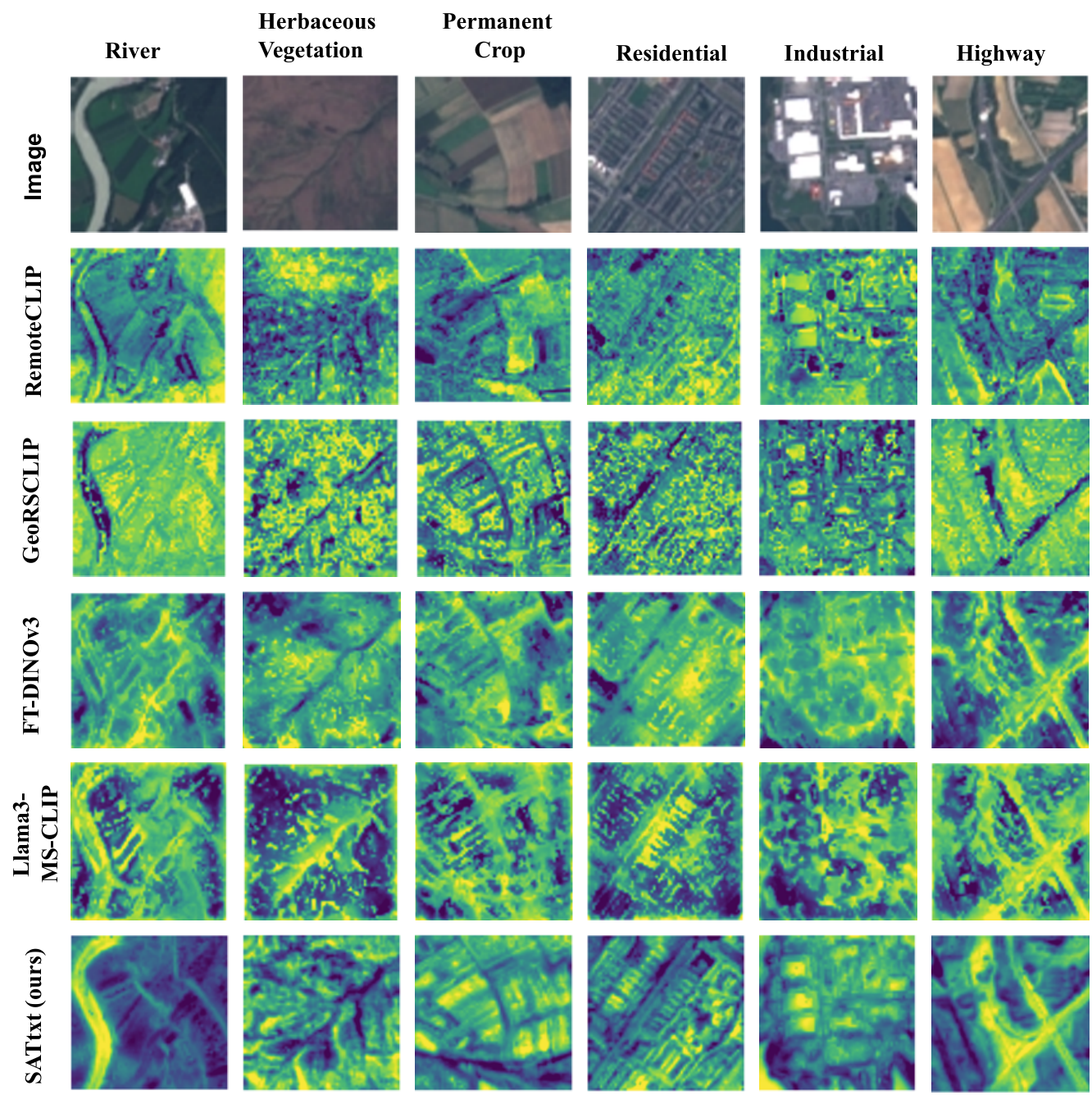}
    \caption{
        Patch-wise image-text similarity maps.
        For each image, we compute cosine similarity between patch-level vision embeddings and the text embedding produced from the prompt
        \emph{``a satellite image of \{class\}''}. 
        Compared to baselines, \gls{method} yields sharper and more contiguous responses that trace class-consistent structures (\eg, linear rivers and highways) and reduce spurious activations in the background.
    }
    \label{fig:feature_map}
\end{figure}

\begin{figure}[h]
    \centering
    \includegraphics[width=1\linewidth]{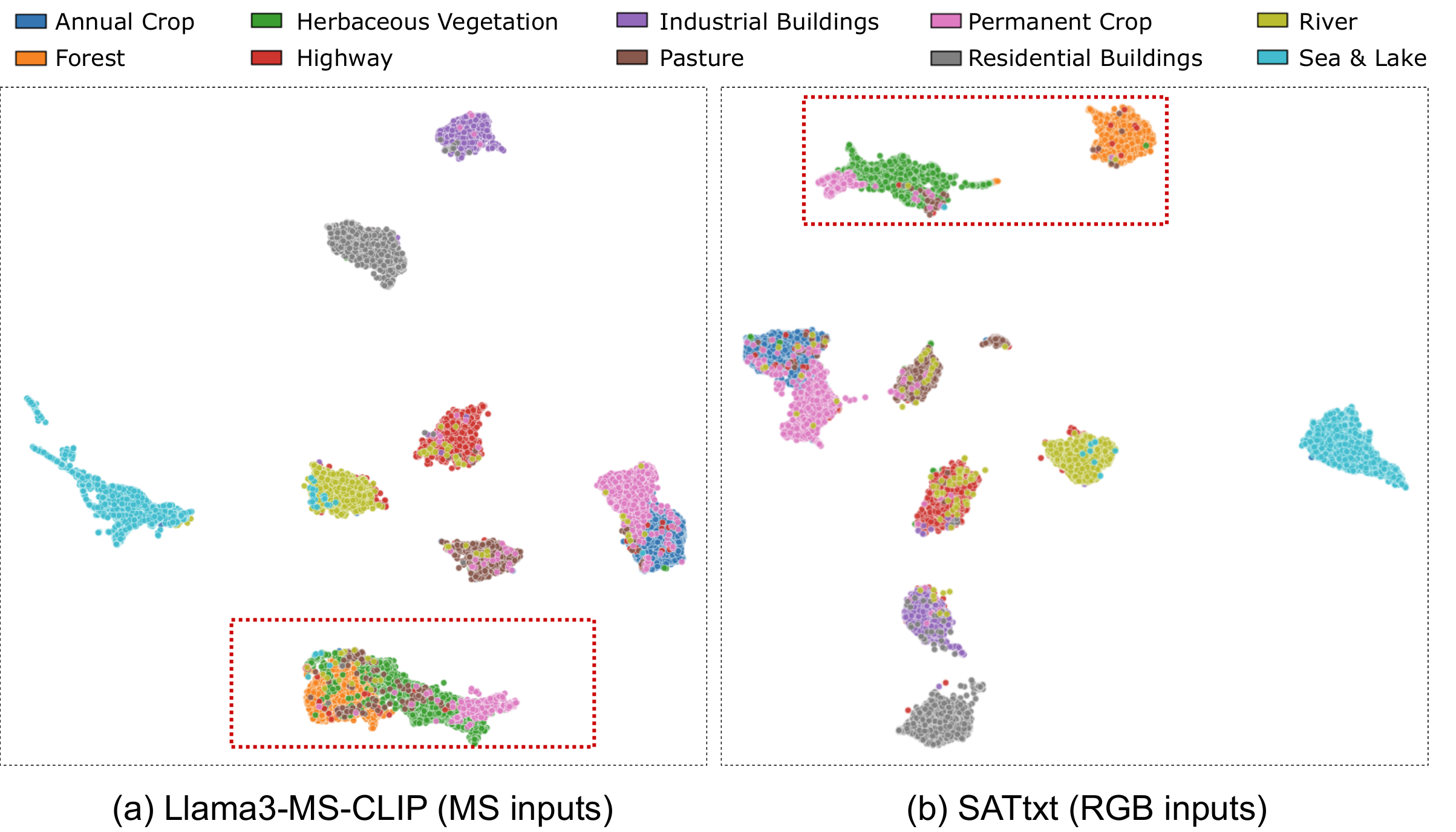}
    \caption{UMAP ($n_{\text{neighbors}} = 50$, $ \text{minDist} = 0$, cosine distance) visualization of feature embeddings for (a) Llama3-MS-CLIP with MS inputs and (b) our \gls{method} with RGB inputs. Points are colored by class. 
     \gls{method} yields more compact intra-class clusters and clearer inter-class separation than Llama3-MS-CLIP.}
    \label{fig:umap}
\end{figure}

Qualitative similarity maps (\cref{fig:feature_map}) align with the quantitative trends. \gls{method} yields sharper, well-localized responses that trace linear structures (\eg rivers) and separate visually similar land-cover types (\eg permanent crops vs.\ herbaceous vegetation). 
Prior models show diffuse or off-target activations. 
\cref{fig:umap} presents UMAP embeddings for the second-best model, Llama3-MS-CLIP, computed with the same configuration reported in that work. Relative to Llama3-MS-CLIP,~\gls{method} forms more compact clusters with clearer separation. The classes ``Herbaceous Vegetation'' and ``Forest'' overlap under Llama3-MS-CLIP, whereas our model separates them. These results indicate that explicit spectral cues and stronger fine-grained text-image alignment are crucial for effective zero-shot transfer.

\noindent
\textbf{Open vocabulary segmentation.}
Consistent with the patch-wise image–text similarity maps employed in open-vocabulary segmentation, \gls{method} demonstrates strong segmentation performance, achieving 31.23 mIoU. This result surpasses the state-of-the-art baseline Llama3-MS-CLIP (28.58 mIoU), which relies on multispectral inputs, as well as FT-DINOv3txt (26.81 mIoU). These comparisons indicate that \gls{method} provides more accurate pixel-level alignment between visual patches and textual concepts under the open-vocabulary setting without segmentation decoder.

\noindent
\textbf{Linear probing.}
To further assess representational richness, we also compare with two recent MIM-based models, SpectralGPT and Terramind.
\cref{tab:linear_probing} shows that domain-adapted contrastive models outperform general-purpose CLIP, with \gls{method} achieving the best overall performance. While results are comparable to Terramind and other MS models when full BigEarthNet supervision is available, \gls{method} yields larger gains in low-data regimes. Notably, the RGB-only \gls{method} outperforms prior RGB and multispectral counterparts, indicating that the spectral-spatial pre-training objective yields embeddings that generalize under RGB-only inputs and transfer well to ForestNet.

\begin{table}
\centering

\caption{Linear-probe on EuroSAT, BigEarthNet (using 10\% and 100\% of training samples), and ForestNet. 
 ``-" denotes an inapplicable dataset.}
\label{tab:linear_probing}

\resizebox{\columnwidth}{!}{%
\begin{tabular}{
    l 
    l 
    c 
    S[table-format=2.2] 
    S[table-format=2.2] 
    S[table-format=2.2] 
    S[table-format=2.2] 
}
\toprule
\multirow{2}{*}{\textbf{\makecell[l]{\textbf{Training}\\\textbf{Approach}}}} & \multirow{2}{*}{\textbf{Model}} & \multirow{2}{*}{\textbf{Input}} & \textbf{EuroSAT} & \multicolumn{2}{c}{\textbf{BigEarthNet}} & \textbf{ForestNet} \\
\cmidrule(lr){5-6}
& & & & \textbf{10\%} & \textbf{100\%} & \\
\midrule
\multirow{2}{*}{MIM} 
& SpectralGPT~\cite{spectralGPT} & MS  & 92.04 & 74.85   & 79.16 & {-} \\
& Terramind~\cite{terramind_iccv_2025}   & MS  & 96.13 & 75.84 & \second{84.68}   & 48.64 \\
\midrule
\multirow{10}{*}{Contrastive} 
& CLIP~\cite{openaiclip}         & RGB & 92.00            & 64.93            & 75.56            & 41.99 \\
& RemoteCLIP~\cite{remoteclip}   & RGB & 96.19            & 64.06            & 75.58            & 42.09 \\
& GeoRSCLIP~\cite{GeoRSCLIP}        & RGB & 95.76            & 77.45            & 81.19   & 46.53 \\
& SkyCLIP~\cite{skyclip}         & RGB & 94.33            & 62.20            & 70.12            & 41.58 \\
& \gls{dinov3txt}~\cite{dinov3}       & RGB & 94.83            & 64.30            & 70.69            & 40.18 \\
& FT-\gls{dinov3txt}                  & RGB & 96.18            & 64.81            & 72.83            & 40.48 \\
& DOFA-CLIP~\cite{dofaclip}      & RGB & \second{96.93}   & 75.37   & 81.37            & \second{49.65} \\
& DOFA-CLIP~\cite{dofaclip}      & MS  & 94.59            & 78.63             & 81.98            & 47.33 \\
& Llama3-MS-CLIP~\cite{ibm_ms_clip} & MS  & 95.00         & \second{78.90}            & 82.44            & {-} \\
\cmidrule(lr){2-7}
& \textbf{\gls{method} (ours)}   & RGB & \best{98.04}     & \best{80.73}     & \best{84.80}     & \best{53.27} \\
\bottomrule
\end{tabular}%
}
\end{table}

\subsection{Ablation Study}
\label{sec:ablation}

\textbf{Setup.} We evaluate three dimensions of ablation. 
First, \emph{component ablation}: starting from a baseline (FT-\gls{dinov3txt}),
we progressively add SRD (Stage~1), replace the CLIP text encoder (CLIPtext) with frozen LLM-based encoders, and finally introduce instruction-augmented prompts. 
Second, \emph{text-side pooling ablation}: fixing the best component configuration, we compare pooling types for forming caption embeddings from LLM token outputs: begin-of-sequence (\texttt{bos}), end-of-sequence (\texttt{eos}), and mean pooling. 
Third, \emph{MS teacher choice}: replacing the SpectralGPT with a less robust model,~\ie, SatMAE~\cite{satmae2022}.

\noindent
\textbf{Component ablation results.}
\cref{tab:components_results} shows that SRD consistently improves the baseline, indicating that spectral distillation injects MS priors into RGB features without requiring MS inputs during later training or inference.
Replacing the CLIPtext with frozen LLM encoders yields further gains, Llama-3.1-8B performs best, likely due to more expressive token-level representations and stronger context modeling.
Instruction-augmented prompting adds additional improvements by embedding task-specific cues for cross-modal alignment.

\begin{figure}
    \centering
    \includegraphics[width=\linewidth]{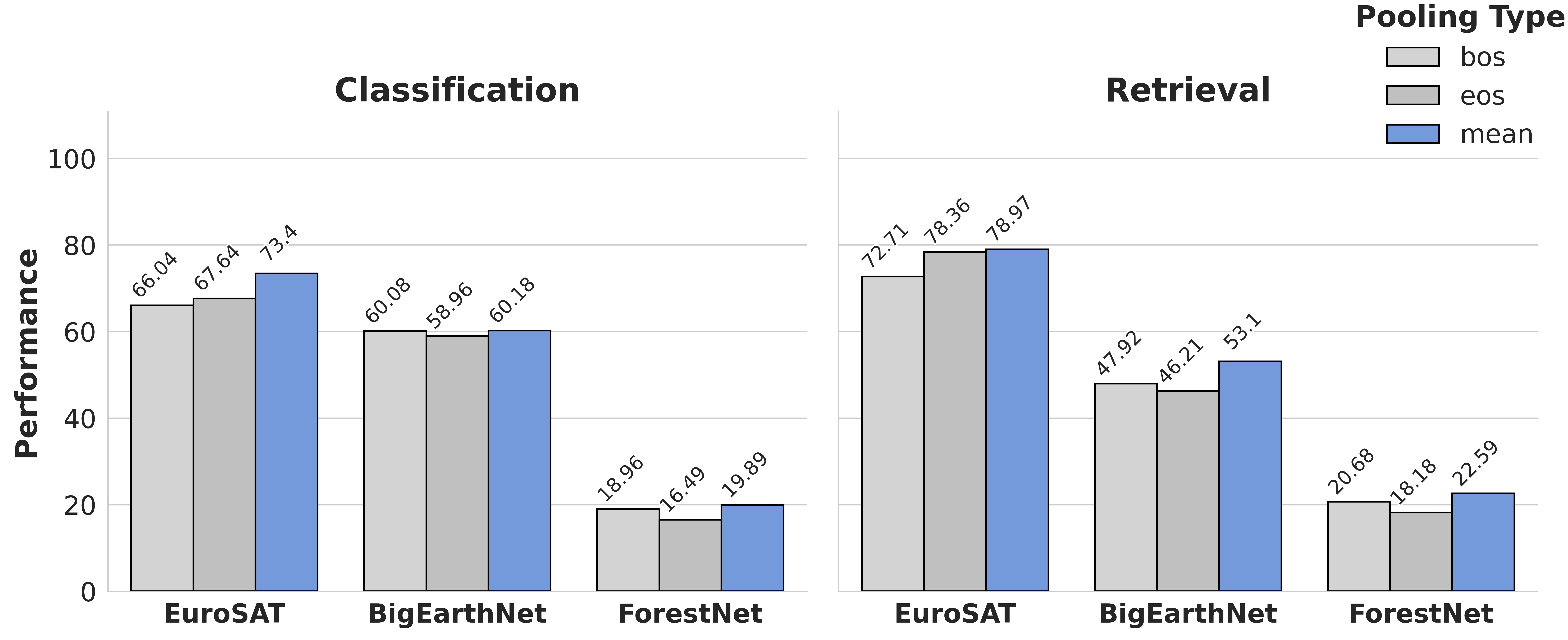}
    \caption{
        Ablation on text-side pooling types.
        Across the datasets and metrics considered, mean pooling yields the most stable and highest performance.
    }
    \label{fig:ablation_pooling}
\end{figure}

\noindent
\textbf{Text-side pooling results.}
\cref{fig:ablation_pooling} illustrates mean pooling outperforms both \texttt{[bos]} and \texttt{[eos]}. 
This suggests that averaging token embeddings leads to a more balanced and stable sentence vector than relying on a single special token. In addition, mean pooling mitigates bias toward token saliency and is more robust to prompt-length variations. This observation aligns with LLM2Vec, which also finds mean pooling to be the most effective aggregation when adapting decoder-only LLMs for embeddings~\cite{llm2vec}.

\begin{table}[t]
\centering
\setlength\tabcolsep{4pt}
\renewcommand{\arraystretch}{0.95}
\caption{
Ablation of model components on three benchmarks.
We incrementally add (i)~\gls{stage1}, (ii) frozen LLM text encoders, and (iii) instruction-augmented prompting.
Every component yields a consistent gain, and the full model (\gls{stage1} + LLM + instruction prompts) achieves the best results on all three benchmarks.
}

\resizebox{\linewidth}{!}{
\begin{tabular}{lcccccc}
\toprule
\multirow{2}{*}{Model} &
\multicolumn{2}{c}{EuroSAT} &
\multicolumn{2}{c}{BigEarthNet} &
\multicolumn{2}{c}{ForestNet} \\
\cmidrule(lr){2-3}\cmidrule(lr){4-5}\cmidrule(lr){6-7}
 & CLS & Retrieval & CLS & Retrieval & CLS & Retrieval \\
\midrule
Baseline (FT-~\gls{dinov3txt}) & 58.6 & 70.6 & 58.1 & 50.9 & 16.7 & 16.3 \\
\midrule
$\text{\gls{method}}_{\text{SRD + CLIPtext}}$ & 65.4 & 74.9 & 58.3 & 51.3 & 17.2 & 19.7 \\
$\text{\gls{method}}_{\text{SRD + Mistral-7B}}$ & 68.2 & 76.2 & 59.1 & 52.3 & 17.5 & 20.1 \\
$\text{\gls{method}}_{\text{SRD + Llama-3.1-8B}}$ & 70.1 & 76.7 & 59.9 & 52.6 & 17.4 & 22.2 \\
$\text{\gls{method}}_{\text{Llama-3.1-8B}}$ + Inst. & 65.3 & 74.7 & 58.3 & 50.2 & 16.9 & 22.0 \\

$\text{\gls{method}}_{\text{SRD + Llama-3.1-8B}}$ + Inst. & \textbf{73.4} & \textbf{79.0} & \textbf{60.2} & \textbf{53.1} & \textbf{17.6} & \textbf{22.6} \\
\bottomrule
\end{tabular}
}
\label{tab:components_results}
\end{table}

\noindent
\textbf{MS teacher choice results.}
\cref{tab:teacher_choice} shows that SATtxt is insensitive to the choice of multi-spectral teacher: SpectralGPT and SatMAE yield closely matched performance and both surpass the strongest baselines overall. 
The small gap between teachers indicates that SRD reliably transfers spectral priors without hinging on a particular teacher, and stronger teachers offer only modest additional gains.

\begin{table}[ht]
\caption{Ablation on MS teacher choice. 
SATtxt exhibits limited sensitivity to the choice of MS teacher.}
\label{tab:teacher_choice}
\centering
\resizebox{\linewidth}{!}{
\begin{tabular}{llrrrrrr}
\toprule
 & & \multicolumn{2}{c}{\textbf{EuroSAT}} & \multicolumn{2}{c}{\textbf{BigEarthNet}} & \multicolumn{2}{c}{\textbf{ForestNet}} \\
\cmidrule(lr){3-4}\cmidrule(lr){5-6}\cmidrule(lr){7-8}
\textbf{Model} & \textbf{Input} & \textbf{CLS} & \textbf{Retrieval} & CLS & \textbf{Retrieval} & CLS & \textbf{Retrieval} \\
\midrule
Llama3-MS-CLIP                  & MS  & 67.86 & 75.26 & 59.63 & 52.42 & -     & -     \\
DOFA-CLIP                       & MS  & 59.04 & 71.54 & 56.58 & 44.52 & 17.02 & 19.55 \\
\midrule
$\text{SATtxt}_\text{SatMAE}$      & RGB & \underline{71.65} & \underline{78.41} & 59.41 & \textbf{53.20} & \underline{17.15} & \underline{20.14} \\
$\text{SATtxt}_\text{SpectralGPT}$ & RGB & \textbf{73.40} & \textbf{78.97} & \textbf{60.18} & \underline{53.10} & \textbf{17.61} & \textbf{22.59} \\
\bottomrule
\end{tabular}
}
\end{table}

\section{Conclusion}
\label{sec:conclusion}
In this paper, we proposed \gls{method}, a spectrum-aware VLFM for satellite imagery that uses only RGB at inference, while transferring MS priors through Spectral Representation Distillation and aligning visual features to text embeddings from instruction-augmented LLMs. 
Across three satellite benchmarks, \gls{method} consistently surpasses strong baselines with the largest gains on the most challenging dataset. 
However, the current design is restricted to optical imagery.
Moreover, although label embeddings are cacheable, relying on a large LLM text encoder increases memory requirements relative to CLIP-style text encoders. Future work will extend the framework to broader sensor coverage, including SAR and thermal imagery.

\section{Acknowledgement}
This work was supported in part by the Australian Government through the Australian Research Council’s Discovery Projects Funding Scheme under Project DP220101634, and by the NVIDIA Academic Grant Program.
{
    \small
    \bibliographystyle{ieeenat_fullname}
    \bibliography{main}
}

\clearpage
\setcounter{page}{1}
\maketitlesupplementary
The Supplementary Material is organized as follows.
~\cref{app:limit_futurework} discusses limitations and future work;
~\cref{app:additional-quant} reports additional quantitative results on downstream tasks;
~\cref{app:additional-qual} provides further qualitative examples;
~\cref{app:baseline_dataset} describes the baselines and downstream datasets;
and ~\cref{app:implementation_details} details the implementation to facilitate reproducibility.

\section{Limitations and Future Work}
\label{app:limit_futurework}
This work is constrained by two limitations: reliance on optical (RGB) imagery and the cost of text encoding at inference. 
Our method studies image-text alignment in optical remote sensing and assumes RGB-only inputs at test time.
Therefore, sensors such as SAR or thermal are not considered, which can limit robustness under cloud cover, haze, or night conditions. 
We chose this setting because RGB is the most commonly available modality. A promising extension is to incorporate additional sensors into the teacher-alignment framework so that the RGB pathway inherits cross-modal priors.
On the efficiency side, using LLMs as the text encoder increases latency and memory during inference.
However, in typical deployments this cost is largely amortized by pre-computing and caching text embeddings for the label vocabulary and prompt sets. 
In addition, the method is compatible with model compression techniques such as quantization, which can reduce runtime and memory footprint while preserving the accuracy.

\section{Additional Quantitative Results}
\label{app:additional-quant}

We report F1 scores for zero-shot classification and for the linear-probe setting.

In \cref{tab:supp_zscls_retrieval}, we compare zero-shot classification F1 across three satellite benchmarks, namely EuroSAT, BigEarthNet, and ForestNet, grouped by training paradigm (generative vs. contrastive) and input modality (RGB vs. MS). Generative models underperform the best contrastive approaches on all datasets. Among contrastive baselines, remote sensing-aware methods such as GeoRSCLIP, SkyCLIP, and DOFA-CLIP reduce the gap to general-purpose vision-language models, and multi-spectral inputs help in some cases, for example DOFA-CLIP with multi-spectral input on ForestNet. Despite using only RGB inputs, \gls{method} with RGB input attains the best results on all three datasets, reaching 73.40 on EuroSAT, 60.18 on BigEarthNet, and 19.89 on ForestNet, and surpassing both RGB and multi-spectral competitors. These results highlight the effectiveness of spectral-aware pre-training even without MS inputs.

\begin{table}[h]
\centering
\caption{Zero-shot classification (F1$\uparrow$) on three satellite benchmarks. ``-" denotes an unsupported task or an inapplicable dataset.}
\label{tab:supp_zscls_retrieval}

\footnotesize
\setlength{\tabcolsep}{3pt}
\resizebox{\columnwidth}{!}{%
\begin{tabular}{
    >{\raggedright\arraybackslash}p{1.8cm}
    l
    c
    S[table-format=2.2]
    S[table-format=2.2]
    S[table-format=2.2]
}
\toprule
\multirow{2}{*}{\makecell[l]{\textbf{Training}\\\textbf{Approach}}} & \multirow{2}{*}{\textbf{Model}} & \multirow{2}{*}{\textbf{Input}} & \multicolumn{3}{c}{\textbf{Zero-shot Classification}} \\
\cmidrule(lr){4-6}
& & & {\textbf{EuroSAT}} & {\textbf{BigEarthNet}} & {\textbf{ForestNet}} \\
\midrule
\multirow{2}{*}{Generative}
& GeoChat & RGB & 30.82 & 14.62 & 6.15 \\
& EarthDial & MS & 57.26 & 31.80 & 8.11 \\
\midrule
\multirow{9}{*}{\makecell[l]{Contrastive\\}}
& CLIP & RGB & 40.35 & 54.85 & 8.30 \\
& RemoteCLIP& RGB & 27.34 & 52.28 & 8.50 \\
& GeoRSCLIP & RGB & 47.04 & 58.80 & 8.27 \\
& SkyCLIP & RGB & 48.50 & 52.88 & 9.73 \\
& \gls{dinov3txt} & RGB & 46.29 & 57.30 & 14.29 \\
& FT-\gls{dinov3txt} & RGB & 53.26 & 58.14 & \second{15.37} \\
& DOFA-CLIP & RGB & 63.73 & 58.96 & 12.33 \\
& DOFA-CLIP & MS & 42.18 & 56.58 & 15.35 \\
& Llama3\text{-}MS\text{-}CLIP & MS & \second{64.27} & \second{59.63} & {-} \\
\cmidrule(lr){2-6}
& \textbf{\gls{method} (ours)} & RGB & \best{73.40} & \best{60.18} & \best{19.89} \\
\bottomrule
\end{tabular}
}
\end{table}

\begin{table}
\centering
\caption{Linear-probe (F1~$\uparrow$) on EuroSAT, BigEarthNet (10\% and 100\%), and ForestNet. “-” denotes an inapplicable dataset.}
\label{tab:supp_linear_probing_f1}

\resizebox{\columnwidth}{!}{%
\begin{tabular}{
    l 
    l 
    c 
    S[table-format=2.2] 
    S[table-format=2.2] 
    S[table-format=2.2] 
    S[table-format=2.2] 
}
\toprule
\multirow{2}{*}{\textbf{\makecell[l]{\textbf{Training}\\\textbf{Approach}}}} &
\multirow{2}{*}{\textbf{Model}} &
\multirow{2}{*}{\textbf{Input}} &
\textbf{EuroSAT} &
\multicolumn{2}{c}{\textbf{BigEarthNet}} &
\textbf{ForestNet} \\
\cmidrule(lr){5-6}
& & & & \textbf{10\%} & \textbf{100\%} & \\
\midrule
\multirow{2}{*}{MIM}
& SpectralGPT     & MS  & 95.96 & 60.70           & 69.93           & {-} \\
& Terramind & MS  & 91.65 & 65.49  & \second{74.08}  & 37.30 \\
\midrule
\multirow{10}{*}{Contrastive}
& CLIP          & RGB & 91.83 & 45.32           & 64.12           & 24.02 \\
& RemoteCLIP    & RGB & 95.02 & 45.40           & 64.24           & 18.35 \\
& GeoRSCLIP         & RGB & 95.76 & 60.71           & 70.65           & 34.74 \\
& SkyCLIP          & RGB & 93.13 & 58.21           & 44.15           & 16.09 \\
& DINOv3txt        & RGB & 94.76 & 26.86           & 45.28           & \second{41.49} \\
& FT-DINOv3txt                   & RGB & 95.46 & 58.31           & 48.64           & 41.13 \\
& DOFA-CLIP       & RGB & \second{96.88} & 59.35 & 65.59           & 12.17 \\
& DOFA-CLIP       & MS  & 94.50 & 66.21           & 69.41           & 13.35 \\
& Llama3-MS-CLIP & MS & 94.85 & \second{68.15}           & 68.04           & {-} \\
\cmidrule(lr){2-7}
& \textbf{\gls{method} (ours)}    & RGB & \best{97.99} & \best{70.43} & \best{74.49} & \best{45.98} \\
\bottomrule
\end{tabular}%
}
\end{table}

\cref{tab:supp_linear_probing_f1} presents linear-probe F1 on EuroSAT, BigEarthNet with 10\% and full supervision, and ForestNet. The comparison includes masked-image-modeling pre-training, represented by SpectralGPT and Terramind, together with contrastive baselines. MIM approaches are competitive, particularly on EuroSAT and BigEarthNet, yet several contrastive models show substantial gains when increasing label fractions.
\gls{method} with RGB input achieves the best performance in every setting, namely 97.99 on EuroSAT, 70.43 and 74.49 on BigEarthNet with 10\% and 100\% labels, and 45.98 on ForestNet, outperforming the strongest multi-spectral and RGB baselines including DOFA-CLIP and Llama3-MS-CLIP. Gains in both low-label and full-label regimes indicate that \gls{method} learns features that are label efficient and linearly separable, and the strong ForestNet result demonstrates robustness real-world forest categories. Overall, the linear-probe trends align with the zero-shot findings, showing that spectral-aware contrastive pre-training yields consistently strong and robust representations across multiple optical satellite datasets.

\begin{figure}[h]
    \centering
    \includegraphics[width=0.9\linewidth]{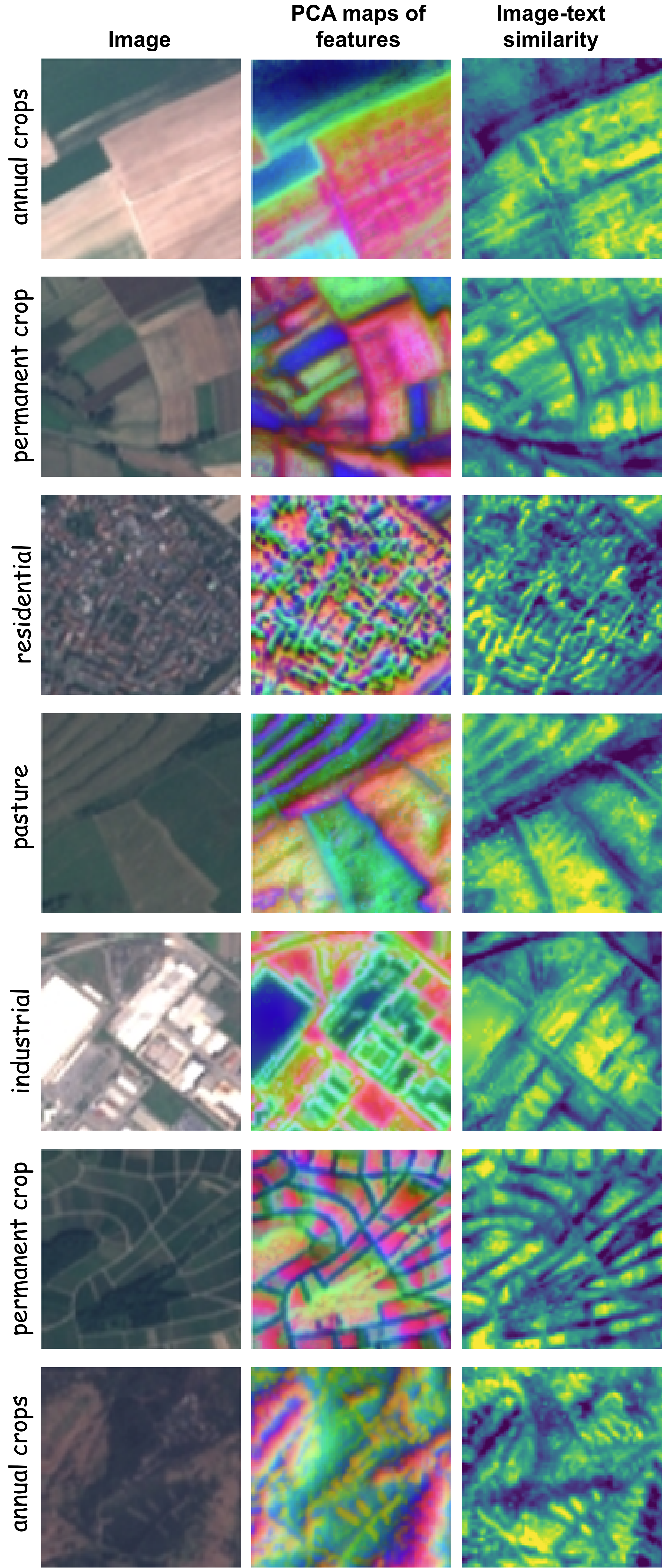}
   \caption{PCA-based feature map visualizations and patch-wise image-text similarity maps with labels.}
    \label{fig:additional_visual}
\end{figure}

\section{Additional Qualitative Results}
\label{app:additional-qual}

We visualize the model’s feature maps following the DINOv3 protocol. Using a patch size of 16, we first resize each image to \(1024\times1024\) pixels, resulting in feature maps of spatial resolution \(64\times64\) at the last layer. We then perform principal component analysis (PCA) on these feature maps and project them onto the first three principal components to obtain pseudo-RGB visualizations, which are bilinearly upsampled back to the original image resolution. In addition, we visualize patch-wise image-text similarity with labels. As shown in \cref{fig:additional_visual}, the SATtxt produces well-structured PCA feature maps and object-focused image-text similarity maps.

\section{Baseline \& Dataset Details}
\label{app:baseline_dataset}
\noindent
\textbf{Baseline.} \cref{tab:supp_model-configs} summarizes the baseline models evaluated in this study. 
We include both general-purpose vision-language models and those specifically adapted for remote sensing imagery. 
All models are configured using their publicly available checkpoints that provide the best performance.

\begin{table}[!h]
\centering
\caption{Model configurations used in our experiments.}
\footnotesize
\setlength{\tabcolsep}{6pt}
\renewcommand{\arraystretch}{1.05}
\begin{tabular}{lcl}
\toprule
\textbf{Model}  & \textbf{Backbone} \\
\midrule
CLIP                & ViT-L-14 \\
RemoteCLIP          & ViT-L-14 \\
GeoRSCLIP          & ViT-B-32 \\
SkyCLIP         & ViT-L-14 \\
Llama3-MS-CLIP      & ViT-B-16 \\
DOFA-CLIP          & ViT-L-14 \\
\gls{dinov3txt}         & ViT-L-16 \\
\gls{method} (ours) & ViT-L-16 \\
\bottomrule
\end{tabular}
\label{tab:supp_model-configs}
\end{table}

\noindent
\textbf{Dataset.} \cref{tab:dataset-details} summarizes the downstream datasets evaluated in this study. 
We consider four publicly available benchmarks spanning two optical satellite sensors: Sentinel-2 and Landsat-8.
For each dataset, we report the number of training/validation/test samples, spectral bands, and classes. Note that BigEarthNet-10\% and BigEarthNet-100\% share the same test split.
EuroSAT includes 13 spectral bands, whereas BigEarthNet includes 12 bands.
For multi-spectral inputs on EuroSAT and BigEarthNet, following Clive~\etal~\cite{ibm_ms_clip}, we exclude bands with mismatched spatial resolutions or redundant information~\cite{satmae2022} and use the 10-band subset \{B02 - B08, B8A, B11, B12\} to ensure fair and consistent comparisons across multi-spectral models. 
To assess cross-sensor generalization, we additionally evaluate on ForestNet~\cite{forestnet}, collected with Landsat-8, which provides 5 spectral bands and 12 classes in a single-label setting.

\begin{table}[!h]
\centering
\caption{Downstream datasets used in this study.}
\footnotesize
\renewcommand{\arraystretch}{1.05}
\resizebox{\columnwidth}{!}{%
\begin{tabular}{lcccc}
\toprule
\textbf{Details} & \textbf{EuroSAT} & \textbf{BigEarthNet-10\%} & \textbf{BigEarthNet-100\%} & \textbf{ForestNet} \\
\midrule
\textbf{\# Training Samples}   & 16,200   & 25,000   & 269,695 & 6,464 \\
\textbf{\# Validation Samples} & 5,400    & 10,000   & 123,723 & 989 \\
\textbf{\# Test Samples}       & 5,400    & 125,866  & 125,866 & 993 \\
\textbf{\# Spectral Bands}     & 13       & 12       & 12      & 5 \\
\textbf{\# Classes}            & 10       & 19       & 19      & 12 \\
\bottomrule
\end{tabular}
}
\label{tab:dataset-details}
\end{table}

\section{Additional Implementation Details}
\label{app:implementation_details}
\begin{figure}
    \centering
    \includegraphics[width=1\linewidth]{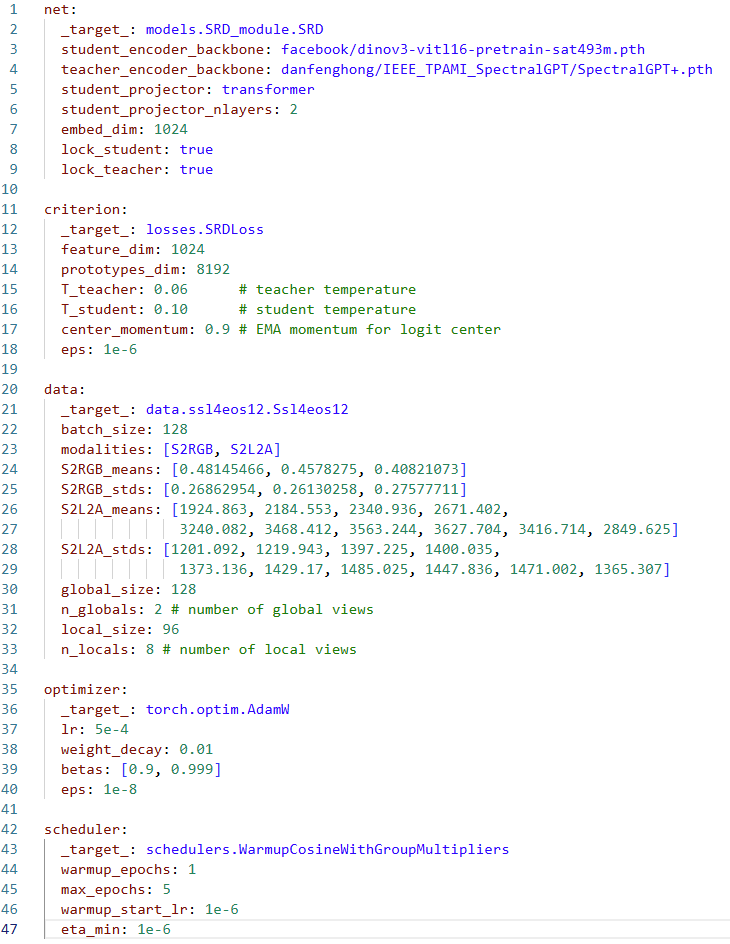}
    \caption{Stage~1 (\gls{stage1}) configuration details.}
    \label{fig:phase_1_implementation_details}
\end{figure}

\begin{figure}
    \centering
    \includegraphics[width=1\linewidth]{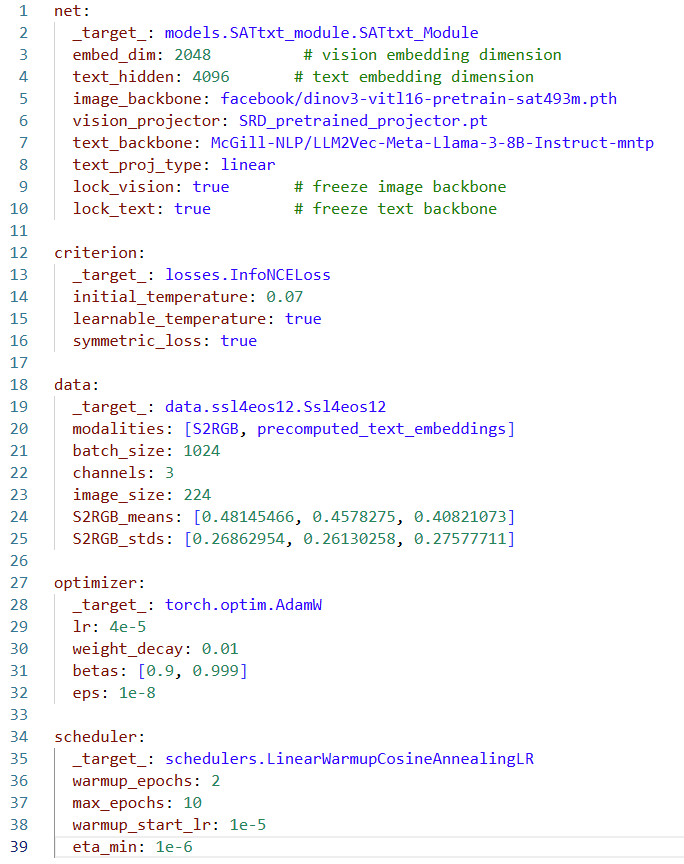}
    \caption{Stage~2 (\gls{stage2}) configuration details.}
    \label{fig:phase_2_implementation_details}
\end{figure}

\textbf{Pre-training Implementation.}
In the first Stage (\gls{stage1}), the multi-spectral (MS) teacher (\ie, SpectralGPT) operates at the fixed $128\times128$ resolution imposed by its pretrained checkpoint.
Accordingly, we set the teacher’s global resolution to $128\times128$ with two views. 
Following DINOv3, the student receives eight local views at $96\times96$. 
We use softmax temperatures of 0.1 for the student and 0.06 for the teacher, with a center momentum of 0.9. 
For data augmentation, we employ multi-crop augmentation that samples two shared random-resized global crops for RGB and MS plus eight RGB-only local crops, applying view-dependent color jitter, grayscale, blur and solarization to RGB while leaving MS geometrically aligned but photometrically clean, followed by per-modality normalization.
Training proceeds for 5 epochs with a batch size of 128 using AdamW, an initial learning rate of $5\times10^{-4}$, and cosine decay. In the second stage (\gls{stage2}), we employ the simple instruction set summarized in~\cref{tab:supp_instruction_list}.
Training runs for 10 epochs with a batch size of 1024 under AdamW optimizer, with an initial learning rate of $4\times10^{-5}$ and cosine decay. 
Complete configuration details for Stage~1 (\gls{stage1}) and Stage~2 (\gls{stage2}) are provided in~\cref{fig:phase_1_implementation_details} and~\cref{fig:phase_2_implementation_details}, respectively.


\textbf{Linear-probe Implementation.}
Unlike zero-shot evaluation, linear-probe results can vary across models because of differences in training hyperparameters and preprocessing. To ensure comparability, we adopt the PANGAEA configuration with minimal preprocessing: resizing, selecting the RGB channels for models that accept RGB inputs, and standard mean-std normalization. All linear probes are trained for 30 epochs with a batch size of 128 using AdamW (learning rate \(1\times10^{-4}\), \(\beta=(0.9, 0.999)\), weight decay \(0.05\)) and a multi-step learning-rate scheduler.

\begin{table}[h!]\centering
\begin{minipage}{0.99\columnwidth}\vspace{0mm}    \centering
\begin{tcolorbox} 
    \centering
    \small
     \hspace{-6mm}
\begin{itemize}[leftmargin=7.5mm]
\setlength{\itemsep}{2pt}
\item     ``Represent this satellite caption to align with its image"
\item    ``Represent this overhead description for image-text retrieval"
\item   ``Remote sensing caption to match its satellite image"
\item    ``Overhead scene description for image-text alignment"
\item    ``Produce a caption representation suitable for visual search over satellite images"

\end{itemize}

\end{tcolorbox}
    
\vspace{-2mm}
\caption{The list of instructions for vision-language alignment.}
\label{tab:supp_instruction_list}
\end{minipage}
\end{table}

\end{document}